\documentclass[final,letterpaper]{scrartcl}
\pdfoutput=1
\usepackage{xcolor}

\usepackage{amsmath}

\usepackage[hyperindex,breaklinks]{hyperref}

\usepackage[utf8]{inputenc}

\usepackage[capitalise,noabbrev]{cleveref}

\usepackage{tikz}
\usetikzlibrary{positioning}

\usepackage[
   natbib,
   style=authoryear-comp,
   citestyle=authoryear-comp,
   bibencoding=utf8,
   useprefix=true,
   maxbibnames=10,
   minbibnames=5,
   maxcitenames=2,
   ]{biblatex}

\DeclareNameFormat{labelname:poss}{
  \nameparts{#1}
  \ifcase\value{uniquename}%
    \usebibmacro{name:family}{\namepartfamily}{\namepartgiven}{\namepartprefix}{\namepartsuffix}%
  \or
    \ifuseprefix
      {\usebibmacro{name:first-last}{\namepartfamily}{\namepartgiveni}{\namepartprefix}{\namepartsuffixi}}
      {\usebibmacro{name:first-last}{\namepartfamily}{\namepartgiveni}{\namepartprefixi}{\namepartsuffixi}}%
  \or
    \usebibmacro{name:first-last}{\namepartfamily}{\namepartgiven}{\namepartprefix}{\namepartsuffix}%
  \fi
  \usebibmacro{name:andothers}%
  \ifnumequal{\value{listcount}}{\value{liststop}}{'s}{}}
\DeclareFieldFormat{shorthand:poss}{%
  \ifnameundef{labelname}{#1's}{#1}}
\DeclareFieldFormat{citetitle:poss}{\mkbibemph{#1}'s}
\DeclareFieldFormat{label:poss}{#1's}
\newrobustcmd*{\posscitealias}{%
  \AtNextCite{%
    \DeclareNameAlias{labelname}{labelname:poss}%
    \DeclareFieldAlias{shorthand}{shorthand:poss}%
    \DeclareFieldAlias{citetitle}{citetitle:poss}%
    \DeclareFieldAlias{label}{label:poss}}}
\newrobustcmd*{\posscite}{%
  \posscitealias%
  \textcite}
\newrobustcmd*{\Posscite}{\bibsentence\posscite}
\newrobustcmd*{\posscites}{%
  \posscitealias%
  \textcites}

\makeatletter
\DeclareCiteCommand{\fullcite}
  {\defcounter{maxnames}{\blx@maxbibnames}%
    \usebibmacro{prenote}}
  {\usedriver
     {\DeclareNameAlias{sortname}{default}}
     {\thefield{entrytype}}}
  {\multicitedelim}
  {\usebibmacro{postnote}}
  \makeatother

{\begin{abstract}\noindent}%
{\end{abstract}}

\providecommand{\paragraph}{}
\let\originalparagraph\paragraph
\renewcommand{\paragraph}[2][.]{\originalparagraph{#2#1}}

\title{AGI Safety Literature Review%
  \thanks{Cite as \fullcite{Everitt2018litrev}.}
  \thanks{We thank Matthew Graves, Kaj Sotala, Ramana Kumar and Daniel
    Eth for comments and suggestions.}
}
\author{Tom Everitt, Gary Lea, Marcus Hutter
  \\\small\texttt{\{tom.everitt, gary.lea, marcus.hutter\}@anu.edu.au}}
\publishers{Australian National University}

\begin{document}
\maketitle

\begin{abstract}
  The development of Artificial General Intelligence (AGI) promises to be a major event.
  Along with its many potential benefits, it also raises serious safety
  concerns (Bostrom, 2014).
  The intention of this paper is to provide an easily accessible and up-to-date
  collection of references for the emerging field of AGI safety.
  A significant number of safety problems for AGI have been identified.
  We list these, and survey recent research on solving them.
  We also cover works on
  how best to think of AGI from the limited knowledge we have today,
  predictions for when AGI will first be created, and what will happen
  after its creation.
  Finally, we review the current public policy on AGI.
\end{abstract}


\tableofcontents
\pagebreak

\section{Introduction}

An Artificial General Intelligence (AGI) is an AI system that equals or
exceeds human intelligence in a wide variety of cognitive tasks.
This is in contrast to today's AI systems that solve only narrow sets of tasks.
Future AGIs may pose significant risks in addition to their many potential
benefits \citep{Bostrom2014}.
The goal of this paper is to survey the literature relevant to these risks and
their prevention.

Why study the safety of AGI before it exists, and before we even know
whether it will ever exist?
There are at least two types of reasons for this.
The first is pragmatic.
If AGI is created, and we do not know how to control it, then the outcome
could be catastrophic \citep{Bostrom2014}.
It is customary to take precautions not only against catastrophes we know
will happen, but also against catastrophes that have only a slight chance
of occurring (for example, a city may decide to build earthquake safe
buildings, even if the probability of an earthquake occurring is fairly low).
As discussed in \cref{sec:predict-agi}, AGI has more than a small probability
of occurring, and it can cause significant catastrophes. 

The second reason is scientific.
Potential future AGIs are theoretically interesting objects,
and the question of how humans can control machines more intelligent
than themselves is philosophically stimulating.
\Cref{sec:understanding-agi} summarizes progress made on understanding
AGI, and \cref{sec:problems-agi,sec:design-agi} consider ways in which
this understanding has helped us to identify problems and generate solutions.

An extensive survey of the AGI safety
literature was previously made by \citet{Sotala2014}.
Since then, the field has grown significantly.
More up-to-date references are provided by this chapter,
and by a number of recent research agendas and problem collections
\citep{Russell2016,Amodei2016,Leike2017gw,Stoica2017,Soares2017,Taylor2016agenda}.
A recent inventory of
AGI projects and their attitudes towards ethics and safety also
contributes to an overview of AGI safety research
and attitudes \citep{Baum2017projectsurvey}.

This paper is structured as follows.
Progress on how to think about yet-to-be-designed
future AGI's is described in the first section (\cref{sec:understanding-agi}).
Based partly on this understanding,
we next survey predictions for when AGI will be created
and what will happen after its creation (\cref{sec:predict-agi}).
We list and discuss identified AGI safety problems (\cref{sec:problems-agi}),
as well as proposals for solving or mitigating them (\cref{sec:design-agi}).
Finally, we review the current public policy on AGI safety issues
(\cref{sec:public-policy}), before making some concluding remarks
(\cref{sec:litrev-conclusions}).

\section{Understanding AGI}
\label{sec:understanding-agi}

A major challenge for AGI safety research is to find the right conceptual
models for plausible AGIs.
This is especially challenging since we can only guess at the
technology, algorithms, and structure that will be used.
Indeed, even if we had the blueprint of an AGI system,
understanding and predicting its behavior might still be hard:
Both its design and its behavior could be highly complex.
Nonetheless, several abstract observations and predictions are
possible to make already at this stage.

\subsection{Defining Intelligence}
\label{sec:intelligence-def}

\citet{Legg2007def} propose a formal definition of intelligence
based on algorithmic information theory and the AIXI theory \citep{Hutter2005}.
They compare it to a large number of previously suggested definitions
\citep{Legg2007list}.
Informally, their definition states that:
\begin{quote}
  ``Intelligence measures an agent’s ability to achieve goals in a wide range of
  environments.''
\end{quote}
The definition is non-anthropomorphic, meaning that it can be applied
equally to humans and artificial agents.
All present-day AIs are less intelligent than humans according to this
definition, as each AI is unable to achieve goals beyond a rather
narrow domain.
These domains can be for example ATARI environments \citep{dqn,a3c,dqnrainbow},
board-games \citep{alphago,alphagozero,alphazero},
car-driving \citep{Bojarski2016,Huval2015}.
However, a trend towards greater generality can be observed, with e.g.\ car
driving being a more general task than Chess, and AlphaZero simultaneously
achieving state of the art performance on several challenging board games
\citep{alphazero}.

Following the Legg-Hutter definition, we may expect that a future,
super-human AGI will be able to achieve more goals in a wider range of
environments than humans.
The most intelligent agent according to this definition is AIXI,
which has been studied both mathematically and empirically;
see
\citet{Everitt2018bc,Leike2016phd,Hutter2012onedecade,Hutter2005} for surveys.
Safety work derived from AIXI is reviewed mostly in \cref{sec:design-agi}.

The Legg-Hutter intelligence definition measures what matters for control.
The more intelligent an agent is, the more control it will have over
aspects of the environment relating to its goals.
If two agents with significantly different Legg-Hutter intelligence
have conflicting goals in a shared environment,
then the more intelligent of the two will typically succeed and the
less intelligent fail.
This points to the risks with increasingly intelligent AGIs:
If their goals are not aligned with ours, then there will likely be a point
where their goals will be achieved to the loss of ours \citep{Russell2016fear}.

\subsection{Orthogonality}
\posscite{Bostrom2012,Bostrom2014} \emph{orthogonality thesis}
states that essentially any level of intelligence is compatible with
any type of goal.
Thus it does not follow, as is sometimes believed, that a highly
intelligent AGI will realize that a simplistic goal
such as creating paperclips or computing decimals of $\pi$ is dumb,
and that it should pursue something more worthwhile such as art or human
happiness.
Relatedly, \citet{Hume1738} argued that
\emph{reason} is the slave of \emph{passion}, and that a passion can never
rationally be derived.
In other words, an AGI will employ its intelligence to achieve its goals,
rather than conclude that its goals are pointless.
Further, if we want an AGI to pursue goals that we approve of, we better
make sure that we design the AGI to pursue such goals:
Beneficial goals will not emerge automatically as the system gets smarter.

\subsection{Convergent Instrumental Goals}
\label{sec:convergent-instrumental-goals}

The orthogonality thesis holds for the \emph{end goals} of the system.
In stark contrast, the \emph{instrumental goals} will often coincide for
many agents and end goals
\citep{Omohundro2007,Omohundro2008aidrives,Bostrom2012}.
Common instrumental goals include:
\begin{itemize}
\item Self-improvement: By improving itself, the agent becomes better
  at achieving its end goal.
\item Goal-preservation and self-preservation:
  By ensuring that future versions of itself pursue
  the same goals, the end goal is more likely to be achieved.
\item Resource acquisition:
  With more resources, the agent will be better at achieving the end goals.
\end{itemize}
Exceptions exist, especially in game-theoretic situations where the
actions of other agents may depend on the agent's goals or other properties
\citep{LaVictoire2014}.
For example, an agent may want to change its goals
so that it always chooses to honor contracts.
This may make it easier for the agent to make deals with other agents.

\subsection{Formalizing AGI}
\label{sec:formalizing-agi}

Bayesian, history-based agents have been used to formalize AGI in the
so-called AIXI-framework
(\citealp{Hutter2005}; also discussed in \cref{sec:intelligence-def}).
Extensions of this framework have been developed for studying
goal alignment \citep{Everitt2018alignment},
multi-agent interaction \citep{Leike2016got},
space-time embeddedness \citep{Orseau2012},
self-modification \citep{Orseau2011,Everitt2016sm},
observation modification \citep{Ring2011},
self-duplication \citep{Orseau2014tele1,Orseau2014tele2},
knowledge seeking \citep{Orseau2014ks},
decision theory \citep{Everitt2015cdtedt},
and others \citep{Everitt2018bc}.

Some aspects of reasoning are swept under the rug by AIXI and
Bayesian optimality.
Importantly, probability theory assumes that agents know all the logical
consequences of their beliefs \citep{Gaifman2004}.
An impressive model of \emph{logical non-omniscience}
has recently been developed by \citet{Garrabrant2016,Garrabrant2017}.
Notably, Garrabrant's theory avoids Gödelian obstacles
for agents reasoning about improved versions of themselves
\citep{Fallenstein2014}.
There is also hope that it can provide the foundation for a decision theory
for logically uncertain events,
such as how to bet on the 50th digit of $\pi$ before calculating it.

\subsection{Alternate Views}

Eric Drexler (private communication, 2017) argues that an AGI does not
need to be an \emph{agent} that plans to achieve a goal.
An increasingly automatized AI research and development process
where more and more of AI development
is being performed by AI tools
can become super-humanly intelligent
without having any agent subcomponent.
Avoiding to implement goal-driven agents that make long-term plans
may avoid some safety concerns.
\citet{Drexler2015} outlines a theoretical idea for how to keep AIs specialized.
Relatedly, \citet{Weinbaum2016} criticize the (rational) agent assumption
underpinning most AGI theory.

However, \citet[Ch.~10]{Bostrom2014} and \citet{Gwern2016}
worry that the incentives for endowing a specialized
\emph{tool AI} with more general capabilities will be too strong.
The more tasks that we outsource to the AI, the more it can help us.
Thus, even if it were possible in theory to construct a safe tool AI,
we may not be able to resist creating an agent AGI, especially if several
competing organizations are developing AI and trying to reap its benefits.
It is also possible that a system of tool AIs obtain agent properties, even if
all of its subcomponents are specialized tool AIs.

\section{Predicting AGI Development}
\label{sec:predict-agi}

Based on historical observations of economical and technological progress,
and on the growing understanding of potential future AGIs described in
\cref{sec:understanding-agi}, predictions have been made both for when the first
AGI will be created, and what will happen once it has been created.

\subsection{When Will AGI Arrive?}

There is an ongoing and somewhat heated debate about when we can expect AGI
to be created, and whether AGI is possible at all or will ever be created.
For example, by extrapolating various technology trends until we can emulate a
human brain, \citet{Kurzweil2005} argues that AGI will be created around 2029.
\citet{Chalmers2010} makes a more careful philosophical analysis of
the brain-emulation argument for AI,
and shows that it defeats and/or avoids counter arguments made by
\citet{Lucas1961,Dreyfus1972,Searle1980,Penrose1994}.
Chalmers is less optimistic about the timing of AGI,
and only predicts that it will happen within this century.

Surveys of when AI researchers estimate that human-level AGI will be created
have been made by
\citet{Baum2011}, \citet{Muller2016}, and \citet{Grace2017}. 
\citet{Baum2011} asked 21 attendees at the 2009 Artificial General Intelligence
conference, and found a median of 2045 for superhuman AGI.
\citet{Muller2016} made a bigger survey of 550 people from the
2012 Philosophy and Theory of AI (PT-AI) and AGI conferences,
the Greek Society for Artificial Intelligence (EETN),
as well as the top 100 most cited authors in artificial intelligence.
The medians for the various groups all fell between 2040 and 2050.
\citet{Grace2017} got 352 responses from NIPS and ICML 2015
presenters on the slightly different question of when AGI will accomplish
all tasks better than human workers, and got a median of 2061.
Interestingly, Asian respondents predicted AGI more than 30 years
sooner than North American respondents, with Europeans in the middle
slightly closer to the Asians than to the North Americans.
It is also worth noting that estimates vary widely, from never
to just a few years into the future.

There are also other indicators of when AGI might arrive.
Algorithmic progress have been tracked by
\citet{Grace2013}, \citet{Eckersley2018}, and \citet{Impacts2018alg},
and the costs of computing have been tracked by
\citet{Impacts2018comp}.
Notably, the computing power available for AI has doubled roughly
every 3-4  months in recent years \citep{Amodei2018}.
A new MIT course on AGI shows that the AGI prospect
is becoming more mainstream \citep{Fridman2018}.
Stanford has a course on AI safety \citep{Sadigh2017}.
\citet{Jilk2017} argues that an AGI must have a conceptual-linguistic faculty in
order to be able to access human knowledge or interact effectively with the
world, and that the development of systems with
conceptual-linguistic ability can be used as an indicator of AGI being near.

\subsection{Will AGI Lead to a Technological Singularity?}

As explained in \cref{sec:convergent-instrumental-goals}, one of the
instrumental goals of almost any AGI will be self-improvement.
The greater the improvement, the likelier the end goals will be achieved.
This can lead to \emph{recursive} self-improvement, where a self-upgraded AGI
is better able to find yet additional upgrades, and so on.
If the pace of this process increases, we may see an
\emph{intelligence explosion} once a critical level of self-improvement
capability has been reached
\citep{Good1966,Vinge1993,Kurzweil2005,Yudkowsky2008,Hutter2012explode,Bostrom2014}.
Already John von Neumann have been quoted calling this intelligence explosion
a \emph{singularity} \citep{Ulam1958}.
Singularity should here not be understood in its strict mathematical sense,
but more loosely as a point where our models break.

Some counter arguments to the singularity have been structured by
\citet{Walsh2016},
who argues that an intelligence explosion is far from inevitable:
\begin{itemize}
\item Intelligence measurement:
  The singularity predicts an increasingly rapid development of intelligence.
  However, it is not quite clear how we should measure intelligence
  \citep{Hutter2012explode}.
  A rate of growth that looks fast or exponential according to one type of
  measurement, may look ordinary or linear according to another measurement
  (say, the log-scale).
\item Fast thinking dog:
  No matter how much we increase the speed at which a dog thinks,
  the dog will never beat a decent human at chess.
  Thus, even if computers keep getting faster, this alone does not entail
  their ever becoming smarter than humans.
\item Anthropocentric:
  Proponents of the singularity often believes that somewhere around
  the human level of intelligence is a critical threshold, after which we
  may see quick recursive self-improvement.
  Why should the human level be special?
\item Meta-intelligence, diminishing returns, limits of intelligence,
  computational complexity:
  It may be hard to do self-improvement or be much smarter than humans
  due to a variety of reasons, such as a fundamental (physical) upper bound
  on intelligence or difficulty of developing machine learning algorithms.
\end{itemize}

These arguments are far from conclusive, however.
In \emph{Life 3.0}, \citet{Tegmark2017} argues that AGI constitutes a third
stage of life.
In the first stage, both hardware and software is evolved (e.g.\ in bacteria).
In the second stage, the hardware is evolved but the software is designed.
The prime example is a human child who goes to school and improves her
knowledge and mental algorithms (i.e.\ her software).
In the third stage of life, both the software and hardware is designed,
as in an AGI.
This may give unprecedented opportunities for quick development,
countering the anthropocentric argument by Walsh.
In relation to the limits of intelligence arguments,
\citet{Bostrom2014} argues that an AGI may think up to a million times
faster than a human.
This would allow it to do more than a millennium of mental work in a day.
Such a speed difference would make it very hard for humans to control the AGI.
Powerful mental representations may also allow an AGI to quickly supersede
human intelligence in quality \citet{Sotala2017}, countering the fast-thinking
dog argument.
The possibility of brain-emulation further undermines the fast-thinking dog argument.
\citet{Yampolskiy2017} also replies to Walsh's arguments.

\posscite{Kurzweil2005} empirical case for the singularity
has been criticized for lack of scientific rigor \citep{Modis2006}.
\citet{Modis2002} argues that a logistic function fits the
data better than an exponential function, and that
logistic extrapolation yields that the rate of complexity growth
in the universe should have peaked around 1990.

In conclusion, there is little consensus on whether and when AGI will be
created, and what will happen after its creation.
Anything else would be highly surprising, given that no similar event
have previously occurred.
Nonetheless, AGI being created within the next few decades and quickly superseding
human intelligence seems like a distinct possibility.

\subsection{Risks Caused by AGI}

A technological singularity induced by an AGI may lead to existential risks
\citep{Bostrom2013,Bostrom2014},
as well as risks of substantial suffering \citep{Sotala2017s-risk}.
However, even if AGI does not lead to a technological singularity, it may still
cause substantial problems, for example through (enabling greater degrees of)
social manipulation, new types of warfare, or shifts in power dynamics
\citep{Sotala2018disjunctive}.
Categorizations of possible scenarios have been proposed by
\citet{Yampolskiy2015pathways,Turchin2018}.

\section{Problems with AGI}
\label{sec:problems-agi}

Several authors and organizations have published research agendas that
identify potential problems with AGI.
\citet{Russell2016} and the Future of Life Institute (FLI) take the broadest view,
covering societal and technical challenges in both the near and the long term future.
\citet{Soares2014agenda,Soares2017} at the Machine Intelligence Research Institute
(MIRI)
focus on the mathematical foundations for AGI, including decision theory and
logical non-omniscience.
Several subsequent agendas and problem collections try to bring the sometimes
``lofty'' AGI problems down to concrete machine learning problems:
\citet{Amodei2016} at OpenAI et al.,
\citet{Leike2017gw} at DeepMind, and
\citet{Taylor2016agenda} also at MIRI.
In the agenda by \citet{Stoica2017} at UC Berkeley,
the connection to AGI has all but vanished.
For brevity, we will refer to the agendas by the organization of the first
author, with MIRI-AF the agent foundations agenda by \citet{Soares2014agenda,Soares2017}
and MIRI-ML the machine learning agenda by \citep{Taylor2016agenda}.
\Cref{fig:agendas} shows some connections between the agendas.
\Cref{fig:agendas} also makes connections to research done by 
other prominent AGI safety institutions:
Oxford Future of Humanity Institute (FHI),
Australian National University (ANU), and
Center for Human-Compatible AI (CHAI).

\begin{figure*}
  \centering
  \tikzset{font=\footnotesize, inner sep=0, outer sep=0}
  \tikzset{heading/.style={font=\bfseries\footnotesize}}
  \tikzset{implicit/.style={font=\bfseries\footnotesize,text=blue}}
  \begin{tikzpicture}[node distance=3mm,y=3.5mm,x=7.5mm]

    \node (miri) at (0, 0) [heading] {MIRI-AF};
    \node (miri-value) [below of=miri] {Value specification};
    \node (miri-error) [below of=miri-value] {Error-Tolerance};
    \node (miri-reliability) [below of=miri-error] {Reliability};

    \node (fli) at (0, -6) [heading] {FLI};
    \node (fli-validity) [below of=fli] {Validity};
    \node (fli-control) [below of=fli-validity] {Control};
    \node (fli-verification) [below of=fli-control] {Verification};
    \node (fli-security) [below of=fli-verification] {Security};

    \node (anu) at (5, 0) [implicit] {ANU};
    \node (anu-rc) [below of=anu] {Reward/data corruption};
    \node (anu-corrigibility) [below of=anu-rc] {Corrigibility};
    \node (anu-sm) [below of=anu-corrigibility] {Self-modification};
    \node (anu-decision) [below of=anu-sm] {Decision theory};

    \node (chai) at (5, -4.5) [implicit] {CHAI};
    \node (chai-reward) [below of=chai] {Reward learning};
    \node (chai-corrigibility) [below of=chai-reward, xshift=2mm] {Corrigibility};

    \node (berkeley) at (5, -7.5) [heading] {UC Berkeley};
    \node (berkeley-secure) [below of=berkeley] {Secure enclaves};
    \node (berkeley-continual) [below of=berkeley-secure] {Continual learning};
    \node (berkeley-robust) [below of=berkeley-continual] {Robust decisions};
    \node (berkeley-explainable) [below of=berkeley-robust] {$\quad$Explainable decisions};
    \node (berkeley-shared) [below of=berkeley-explainable] {Shared learning};
    \node (berkeley-architectures) [below of=berkeley-shared] {AI architectures};
    \node (berkeley-adversarial) [below of=berkeley-architectures] {Adversarial learning};
        
    \node (fhi) at (10, 0) [implicit] {FHI};
    \node (fhi-reward) [below of=fhi] {Reward learning};
    \node (fhi-interruptibility) [below of=fhi-reward] {Interruptibility};
    \node (fhi-corruption) [below of=fhi-interruptibility] {Data corruption};
    \node (fhi-side) [below of=fhi-corruption] {Side effects};
    \node (fhi-oracles) [below of=fhi-side] {Oracles};
    \node (fhi-distillation) [below of=fhi-oracles] {Distillation};
    \node (fhi-exploration) [below of=fhi-distillation] {Safe exploration};
    
    \node (miri2) at (10, -7) [heading] {MIRI-ML};
    \node (miri2-ambiguity) [below of=miri2] {Ambiguity detection};
    \node (miri2-imitation) [below of=miri2-ambiguity] {Human imitation};
    \node (miri2-oversight) [below of=miri2-imitation] {Informed oversight};
    \node (miri2-goals) [below of=miri2-oversight] {Generalizable goals};
    \node (miri2-concepts) [below of=miri2-goals] {Conservative concepts};
    \node (miri2-impact) [below of=miri2-concepts] {Impact measures};
    \node (miri2-mild) [below of=miri2-impact] {Mild optimization};
    \node (miri2-instrumental) [below of=miri2-mild] {Instrumental incentives};

    \node (deepmind) at (15, 0) [heading] {DeepMind};
    \node (deepmind-absent) [below of=deepmind] {Absent supervisor};
    \node (deepmind-interruptibility) [below of=deepmind-absent] {Interruptibility};
    \node (deepmind-gaming) [below of=deepmind-interruptibility] {Reward gaming};
    \node (deepmind-modification) [below of=deepmind-gaming] {Self-modification};
    \node (deepmind-side) [below of=deepmind-modification] {Negative side effects};
    \node (deepmind-exploration) [below of=deepmind-side] {Safe exploration};
    \node (deepmind-shift) [below of=deepmind-exploration] {Distributional shift};
    \node (deepmind-adversaries) [below of=deepmind-shift] {Adversaries};

    \node (openai) at (15, -8) [heading] {OpenAI et al.};
    \node (openai-exploration) [below of=openai] {Safe exploration};
    \node (openai-corruption) [below of=openai-exploration] {Reward corruption};
    \node (openai-oversight) [below of=openai-corruption] {Scalable oversight};
    \node (openai-shift) [below of=openai-oversight] {Distributional shift};
    \node (openai-side) [below of=openai-shift] {Negative side effects};
    
    \path
    (miri-value.west) edge (fli-validity.west)
    (miri-reliability.west) edge (fli-validity.west)
    (miri-error.west) edge[bend right] (fli-control.west)

    (fli-security.east) edge (berkeley-secure.west)
    (fli-control.east) edge (chai-corrigibility.west)

    (anu-rc) edge (miri-value)
    (anu-rc.east) edge (fhi-corruption.west)
    (anu-corrigibility) edge (miri-error)
    (anu-sm) edge (miri-reliability)
    (anu-decision) edge (miri-reliability)

    (chai-reward.west) edge (miri-value.east)
    (chai-reward.east) edge (miri2-goals.west)
    (chai-reward.east) edge (miri2-imitation.west)

    (berkeley-robust.east) edge (miri2-ambiguity.west)
    (berkeley-explainable.east) edge (miri2-oversight.west)

    (fhi-reward.west) edge (chai-reward.east)
    (fhi-interruptibility.west) edge (chai-corrigibility.east)
    (fhi-interruptibility.east) edge (deepmind-interruptibility.west)
    (fhi-exploration.east) edge (deepmind-exploration.west)
    (fhi-corruption.east) edge (deepmind-gaming.west)
    (fhi-side.east) edge (deepmind-side.west)

    (miri2-impact.east) edge (openai-side.west)
    (miri2-mild.east) edge (openai-side.west)
    (miri2-instrumental.east) edge (openai-side.west)
    (miri2-ambiguity.east) edge (deepmind-shift.west)

    (openai-exploration.west) edge (fhi-exploration.east)
    
    (openai-shift.west) edge (miri2-ambiguity.east)
    (openai-oversight.east) edge[bend right=30] (deepmind-absent.east)
    (openai-corruption.east) edge[bend right=35] (deepmind-gaming.east)
    (openai-side.east) edge[bend right=20] (deepmind-side.east)
    ;

    \node (adv-help1) at (8,-14.5) {.};
    \node (adv-help2) at (16,-14.5) {.};
    \path
    (berkeley-adversarial.east) edge[in=180, out=-5] (adv-help1.center)
    (adv-help1.center) edge (adv-help2.center)
    (adv-help2.center) edge[out=0, in=0] (deepmind-adversaries.east);
    
  \end{tikzpicture}
  \caption[AGI Safety Problems]{Connections between problems stated in different AGI safety
    research agendas (for ANU, CHAI, and FHI, the agendas are inferred from
    their recent publications).}
  \label{fig:agendas}
\end{figure*}

Some clusters of problems appear in multiple research agendas:
\begin{itemize}
\item Value specification:
  How do we get an AGI to work towards the right goals?
  MIRI calls this value specification.
  \citet{Bostrom2014} discusses this problem at length, arguing that it is much
  harder than one might naively think.
  \citet{Davis2015} criticizes Bostrom's argument, and \citet{Bensinger2015}
  defends Bostrom against Davis' criticism.
  Reward corruption, reward gaming, and negative side effects are subproblems
  of value specification highlighted in the DeepMind and OpenAI agendas.
\item Reliability:
  How can we make an agent that keeps pursuing the goals we have designed it
  with?
  This is called \emph{highly reliable agent design} by MIRI,
  involving decision theory and logical omniscience.
  DeepMind considers this the self-modification subproblem.
\item Corrigibility:
  If we get something wrong in the design or construction of an agent,
  will the agent cooperate in us trying to fix it?
  This is called error-tolerant design by MIRI-AF and
  \emph{corrigibility} by \citet{Soares2015cor}.
  The problem is connected to safe interruptibility as considered by DeepMind.
\item Security:
  How to design AGIs that are robust to adversaries and adversarial environments?
  This involves building sandboxed AGI protected from adversaries (Berkeley),
  and agents that are robust to adversarial inputs (Berkeley, DeepMind).
\item Safe learning:
  AGIs should avoid making fatal mistakes during the learning phase.
  Subproblems include safe exploration and distributional shift (DeepMind, OpenAI),
  and continual learning (Berkeley).
\item Intelligibility:
  How can we build agent's whose decisions we can understand?
  Connects explainable decisions (Berkeley) and informed oversight (MIRI).
  DeepMind is also working on these issues, see \cref{sec:intelligibility} below.
\item Societal consequences:
  AGI will have substantial legal, economic, political, and military consequences.
  Only the FLI agenda is broad enough to cover these issues,
  though many of the mentioned organizations evidently care about the issue
  \citep{Brundage2018,DeepMind2017}.
\end{itemize}

There are also a range of less obvious problems, which have received
comparatively less attention:
\begin{itemize}
\item
  Subagents:
  An AGI may decide to create \emph{subagents} to help it with its task
  \citep{Soares2015cor,Orseau2014tele1,Orseau2014tele2}.
  These agents may for example be copies of the original
  agent's source code running on additional machines.
  Subagents constitute a safety concern, because even if the
  original agent is successfully shut down, these subagents may not get the message.
  If the subagents in turn create subsubagents, they may spread like a viral disease.
\item
  Malign belief distributions:
  \citet{Christiano2016} argues that the \emph{universal distribution} $M$
  \citep{Solomonoff1964a,Solomonoff1964b,Solomonoff1978,Hutter2005}
  is malign.
  The argument is somewhat intricate, and is based on the idea that a hypothesis
  about the world often includes simulations of other agents, and that these
  agents may have an incentive to influence anyone making decisions based on the
  distribution.
  While it is unclear to what extent this type of problem would affect
  any practical agent, it bears some semblance to aggressive \emph{memes},
  which do cause problems for human reasoning \citep{Dennett1990}.
\item
  Physicalistic decision making:
  The \emph{rational agent} framework is pervasive in the study of artificial
  intelligence.
  It typically assumes that a well-delineated entity interacts with an environment
  through action and observation channels.
  This is not a realistic assumption for \emph{physicalistic} agents such as
  robots
  that are part of the world they interact with \citep{Soares2014agenda,Soares2017}.
\item 
  Multi-agent systems:
  An artificial intelligence may be copied and distributed, allowing
  instances of it to interact with the world in parallel.
  This can significantly boost learning, but undermines the concept of a
  single agent interacting with the world.
\item
  Meta-cognition:
  Agents that reason about their own computational resources and
  \emph{logically uncertain} events can encounter strange paradoxes due to
  Gödelian limitations \citep{Fallenstein2015,Soares2014agenda,Soares2017}
  and
  shortcomings of probability theory
  \citep{Soares2015logical-uncertainty,Soares2014agenda,Soares2017}.
  They may also be \emph{reflectively unstable}, preferring to change
  the principles by which they select actions \citep{Arbital2018}.
\end{itemize}
While these problems may seem esoteric, a security mindset
\citep{Yudkowsky2017security-mindset}
dictates that we should not only protect ourselves from things that can clearly go
wrong, but also against anything that is not guaranteed to go right.
Indeed, unforeseen errors often cause the biggest risks.
For this reason, the biggest safety problem may be one that we have not thought of yet -- not
because it would necessarily be hard to solve, but because in our ignorance
we fail to adopt measures to mitigate the problem.




\section{Design Ideas for Safe AGI}
\label{sec:design-agi}

We next look at some ideas for creating safe AGI.
There is not always a clear line distinguishing ideas for safe AGI from
other AI developments.
Many works contribute to both simultaneously.

\subsection{Value Specification}
\label{sec:design-vl}

\paragraph{RL and misalignment}
Reinforcement learning (RL) \citep{Sutton1998} is currently the most promising
framework for developing intelligent agents and AGI.
Combined with Deep Learning, it has seen some remarkable recent successes,
especially in playing board games \citep{alphago,alphagozero,alphazero}
and computer games \citep{dqn,a3c,dqnrainbow}.

Aligning the goals of an RL agent with the goals of its human supervisor
comprises significant challenges, however~\citep{Everitt2018alignment}.
These challenges include correct specification of the reward function,
and avoiding that the agent takes shortcuts in optimizing it.
Such shortcuts include the agent corrupting the observations on which the reward
function evaluates performance, modifying the reward function to give more
reward, hijacking the reward signal or the memory location of the reward,
and, in the case of an interactively learned reward function, corrupting the
data training the reward function.
\citet{Everitt2018alignment}
categorize misalignment problems in RL, and suggest a number of techniques for
managing the various sources of misalignment.
The rest of this subsection reviews other work that has been done on designing
agents with correctly specified values.

\paragraph{Learning a reward function from actions and preferences}
One of the main challenges in scaling RL to the real world includes designing
the reward function.
This is particularly critical for AGI, as a poorly designed reward function
would likely lead to a misaligned agent.
As an example of misalignment,
\citet{Clark2016} found that their boat racing agent preferred
going in circles and crashing into obstacles instead of winning the race,
due to a subtly misspecified reward function.
\citet{Lehman2018,Gwern2011,Irpan2018} have many more examples.
Analogous failures in AGIs could cause severe catastrophes.
The DeepMind problem collection calls this a \emph{reward gaming} problem.
One potential way around the problem of gameable reward functions
is to let the agent learn the reward function.
This lets designers offload some of the design work to powerful machine
learning techniques.
%

Inverse reinforcement learning (IRL) \citep{Ng2000,Ziebart2008,Choi2011}
is a framework for learning a reward function from the actions of an expert,
often a human demonstrator.
In one famous example, \citet{Abbeel2007} taught an agent acrobatic helicopter flight
by observing the actions of a human pilot.
Impressively, the agent ultimately became better at flying
than the pilot it observed.
However, a learned reward function cannot be better than the data that
trained it.
If all training happens before the agent is launched into the environment, then the
data may not properly describe situations that the agent reaches far into its
lifetime (a so-called \emph{distributional shift} problem; \citealp{Amodei2016}).
For this reason, interactive training of the reward function may be preferable,
as it allows the training data to adapt to any new situation the agent may encounter.

Cooperative inverse reinforcement learning (CIRL) is a generalization of IRL
that lets the expert and the agent act simultaneously in the same environment,
with the agent interactively learning the expert's preferences \citep{Hadfield-Menell2016cirl}.
Among other things, this allows the expert to take demonstrative actions
that are suboptimal according to his or her reward function but
more informative to the agent, without the agent being led to infer an
incorrect reward function.
The CIRL framework can be used to build agents that avoid interpreting reward functions
overly literally, thus avoiding some misalignment problems with RL
\citep{Hadfield-Menell2017ird}.

A reward functions can also be learned from a human rating short
video clips of (partial) agent trajectories against each
other \citep{Christiano2017hp}.
For example, if the human consistently rates scenarios where the agent falls
off a cliff lower than other scenarios, then the learned reward function
will assign a low reward to falling off a cliff.
Using this technique, a non-expert human can
teach an agent complex behaviors that would have been difficult
to directly program a reward function for.
\citet{Warnell2017} use a related approach, needing only 15 minutes of human
feedback to teach the agent the ATARI game Bowling.
In order to scale this method to more complex tasks where evaluation
is non-trivial, \citet{Irving2018} propose letting two systems debate
which option is better, highlighting flaws in each others suggestions
and arguments.
Ideally, following the debate will significantly boost the human's ability to
make an informed evaluation.

On a fundamental level, learning from actions and learning from preferences
is not widely different.
Roughly, a choice of action $a$ over action $b$ can be interpreted as a
preference for the future trajectories resulting from action $a$ over the
trajectories resulting from action $b$.
However, a few notable differences can still be observed.
First, at least in \posscite{Christiano2017hp} framework, preferences always
apply to \emph{past} events.
In contrast, an action in the CIRL framework typically gives information about which
\emph{future} events the human prefers.
A drawback is that in order for the action to carry information
about future events, the action must be chosen (somewhat) rationally.
Humans do not always act rationally;
indeed, we exhibit several systematic biases \citep{Kahneman2011}.
A naive application of (C)IRL therefore runs the risk of inferring
an incorrect reward function.
To address this, \citet{Evans2016} develop a method for learning the reward function of
agents exhibiting some human-like irrationalities.
Without assumptions on the type of irrationality the expert exhibits,
nothing can be learned about the reward function
\citep{Armstrong2017impossibility}.
In comparison, learning from preferences seems to require weaker rationality
assumptions on the human's part, as correctly stating ones preferences
may be easier than acting rationally.

Yet another approach to learning a reward function is to
learn it from stories \citep{Riedl2016}.

\paragraph{Approval-directed agents}
In a series of blog posts, \citet{Christiano2014}
suggests that AGIs should be designed to maximize approval for their actions
rather than trying to reach some goal.
He argues that approval-directed systems have
many of the same benefits of goal-directed systems while avoiding some of their
worst pitfalls.
\citet{Christiano2015,Cotra2018} outline a method for how
approval-directed agents can be chained together in a hierarchy,
boosting the accuracy of the approvals of the human at the top of the
chain.

\paragraph{Reward corruption}

Reinforcement learning AGIs may hijack their reward signal and feed themselves
maximal reward \citep{Ring2011}.
Interestingly, model-based agents with preprogrammed reward functions are
much less prone to this behavior \citep{Everitt2016sm,Hibbard2012}.
However, if the reward function is learned online as discussed above,
it opens up the possibility of \emph{reward learning corruption}.
An AGI may be tempted to influence the data training its reward function
so it points towards simple-to-optimize reward functions rather than harder ones
\citep{Armstrong2015}.
\citet{Everitt2017rc}
show that the type of data the agent receives matter
for reward learning corruption.
In particular, if the reward data can be cross-checked between multiple sources,
then the reward corruption incentive diminishes drastically.
\citeauthor{Everitt2017rc} also evaluate a few different approaches to reward
learning, finding that the \emph{human action}-data provided in CIRL is much
safer than the \emph{reward}-data provided in standard RL, but that CIRL is not
without worrying failure modes.

\paragraph{Side effects}

An AGI that becomes overly good at optimizing a goal or reward function that
does not fully capture all human values, may cause significant negative side
effects \citep{Yudkowsky2009fragile}.
The \emph{paperclip maximizer} that turns the earth and all humans into
paperclips is an often used example \citep{Bostrom2014}, now available as a
highly addictive computer game \citep{Lantz2017}.
Less extreme examples include companies that optimize profits and cause
pollution and other externalities as negative side effects.

The most serious side effects seem to occur when a target function is optimized
in the extreme (such as turning the earth into paperclips).
Quantilization can avoid over-optimization under some assumptions
\citep{Taylor2016quant,Everitt2017rc}.
Another more specific method is to ``regularize'' reward by the impact the
policy is causing \citep{Armstrong2017impact}.
How to measure impact remains a major open question, however.

\paragraph{Morality}

Unfortunately, there is little consensus about which moral theory is the right
one \citep{Brundage2014}.
The machine learning approaches discussed above may partially sidestep the
need to commit to any particular moral theory.
However, the choice of data set and  learning algorithm may
influence which theory is converged upon \citep{Bogosian2017}.
This justifies a discussion about the choice of moral theory for AGIs.

\citet{Sotala2016humanvalues} proposes a moral theory based on evolutionary
psychology and reinforcement learning, compatible with social
intuitionism~\citep{Haidt2001}.
Other suggestions include game-theory \citep{Letchford2008}
and machine learning \citep{Conitzer2017,Shaw2018}.
For example, \citet{Kleiman-Weiner2017} use hierarchical Bayesian inference
to infer a moral theory from actions, and \citet{Abel2016} suggest
that POMDPs can be used as a formal framework for machine ethics.
Virtue ethics has recently re-emerged as a third contender to
consequentialist (utilitarian) and rule-based (deontological) ethics.
\citet{Murray2017stoic} makes a case for building moral AI systems based
on Stoic virtues rather then consequentialist reward-maximization.
Rejecting all particular choices, \citet{Bogosian2017} instead argues that
a better option would be to build an AGI that remains
uncertain about which moral theory is correct,
since morally uncertain agents typically avoid events that are extremely bad according
to any possibly correct moral theory.
Relatedly, \citet{Vamplew2018} discuss a maximin approach over a class
of utility functions or moral theories.
The class of possibly correct moral theories remains to be specified,
however.
More background is provided in the book on machine ethics by
\citet{Wallach2008}.

\paragraph{Connections to economics}
The goal alignment problem has several connections to the economics literature.
It may be seen as an instance of Goodhart's law
\citep{Goodhart1975}, which roughly states that any measure of performance
ceases to be a good measure once it is optimized for.
\citet{Manheim2018} categorize instances of Goodhart's law.
It may also be seen as a principal-agent problem:
The connections have been fleshed out by
\citet{Hadfield-Menell2018contracting}.

\paragraph{Human-inspired designs}
Most of the above discussion concerns value specification by using machine
learning techniques on behavioral human data.
Some authors suggest that we should instead or additionally use the human brain
directly.
\citet{Eth2017} suggest that whole brain emulation is likely the
best path to safe AGI, especially if we emulate the brain of particularly moral
humans.
On a similar note,
\citet{Jilk2017neuromorphic} suggest that \emph{neuromorphic} AGI architectures
that are inspired by the human brain are preferable to other architectures,
as they permit us to use our extensive knowledge about human drives to
anticipate dangers and to develop a good training program for the agent.

These suggestions are controversial.
\citet[Ch.~14]{Bostrom2014} and \citet{Yudkowsky2008xrisk} give several reasons
why they would be less safe.
For example, an emulated human brain may not be as moral as its physical
instance, as its environment and abilities will be rather different.
A neuromorphic AGI is potentially even less safe, as the safe guards of an exact
copy of a human brain may be gone.
Countering this objection, \citet{Jilk2017neuromorphic} argue that it will
likely be hard to prove properties about any type of AGI.
It is an open question whether any of the promising works we have reviewed above
will lead to mathematical guarantees surpassing our potential trust in a
brain emulation or a neuromorphic AGI.


Another way to use insights from the brain is put forth by \citet{Sarma2017}.
Human values are to a large extent based on neural circuitry shared among all
mammals.
These mammalian values may be used as a prior for
machine learning techniques that can then infer the details from human behavior.
This avoids the problems of brain emulation and neuromorphic AGI, while still
benefiting from insights about the human brain.
As a first step, they suggest systematically verifying relevant
neuroscience results \citep{Sarma2017a}.

\subsection{Reliability}

\paragraph{Self-modification}

Even if the reward function is correctly specified, an AGI may still
be able to corrupt either the reward function itself or the data feeding it.
This can happen either intentionally if such changes can give the agent
more reward, or accidentally as a side effect of the agent trying to
improve itself (\cref{sec:convergent-instrumental-goals}).

A utility self-preservation argument going back to at least \citet{Schmidhuber2007} and
\citet{Omohundro2008aidrives} says that agents should not want to change their
utility functions, as that will reduce the utility generated by their
future selves, as measured by the current utility function.
\citet{Everitt2016sm} formalize this argument,
showing that it holds under three non-trivial assumptions:
(1) The agent needs to be model-based, and evaluate future scenarios
according to its current utility function;
(2) the agent needs to be able to predict how self-modifications affect
its future policy; and
(3) the reward function itself must not endorse self-modifications.
In RL \citep{Sutton1998},
model-free agents violate the first assumption,
off-policy agents such as Q-learning violate the second,
and the third assumption may fail especially in learned reward/utility functions
(\cref{sec:design-vl}).
\citet{Hibbard2012} and \citet{Orseau2011} also study the utility
self-preservation argument.

\paragraph{Decision theory}
Agents that are embedded in their environment, such as robots, introduce
subtleties in how expected utility should be calculated.
The most established decision theories for embedded agents are causal decision
theory and evidential decision theory, formalized in a sequential setting by
\citet{Everitt2015cdtedt}.
Both causal and evidential decision theory have weaknesses, however, as both
seem to prescribe the ``wrong'' decision in some situations.
Functional decision theory consolidates years of research about decision theory
for embedded agents, and appears to avoid all known weaknesses of both causal
and evidential decision theory \citep{Yudkowsky2017fdt}.

\subsection{Corrigibility}

By default, agents may resist shutdown and modifications due to the
self-preservation drives discussed in \cref{sec:convergent-instrumental-goals}.
Three rather different approaches have been developed to counter the
self-preservation drives.

\paragraph{Indifference}
By adding a term or otherwise modifying the reward function,
the agent can be made indifferent between some choices of future events,
for example shutdown or software corrections \citep{Armstrong2017indiff}.
For example, this technique can be used to construct variants of
popular RL algorithms that do not learn to prevent interruptions \citep{Orseau2016}.

\paragraph{Ignorance}
Another option is to construct agents that behave as if a certain event
(such as shutdown or software modification) was certain not to happen
\citep{Everitt2016sm}.
For example, off-policy agents such as Q-learning behave as if they will
always act optimally in the future, thereby effectively disregard
the possibility that their software or policy be changed in the future.
\citet{Armstrong2017indiff} show that ignorance is equivalent to indifference
in a certain sense.

\paragraph{Uncertainty}
In the CIRL framework \citep{Hadfield-Menell2016cirl},
agents are uncertain about their reward function, and learn about the
reward function through interaction with a human expert.
Under some assumptions on the human's rationality and the agent's level
of uncertainty, this leads to naturally corrigible agents
\citep{Hadfield-Menell2016osg,Wangberg2017osg}.
Essentially, the agent will interpret the human's act of shutting them down
as evidence that being turned off has higher reward than remaining turned on.
In some cases where the human is likely to make suboptimal choices, the
agent may decide to ignore a shut down command.
There has been some debate about whether this is a feature \citep{Milli2017}
or a bug \citep{Carey2017osg}.

\paragraph{Continuous testing}
\citet{Arnold2018} argue that an essential component of corrigibility is to
detect misbehavior as early as possible.
Otherwise, significant harm may be caused without available corrigibility
equipment having been put to use.
They propose an ethical testing framework that continually monitors the agent's
behavior on simulated ethics tests.

\subsection{Security}

\paragraph{Adversarial counterexamples}
Deep Learning \citep[e.g.\ ][]{Goodfellow2016book} is a highly versatile tool for
machine learning, and a likely building block for future AGIs.
Unfortunately, it has been observed that small perturbations of inputs can
cause severe misclassification errors
\citep{Szegedy2013,Goodfellow2014,Evtimov2017,Athalye2017}.

In a recent breakthrough, \citet{Katz2017} extend the Simplex algorithm
to neural networks with rectified linear units (Relus).
\citeauthor{Katz2017} call the extended algorithm ReluPlex,
and use it to successfully verify the behavior of neural networks with
300 Relu nodes in 8 layers.
They gain insight into the networks' behaviors in certain important regions,
as well as the sensitivity to adversarial perturbations.

\subsection{Intelligibility}
\label{sec:intelligibility}

While it is infamously hard to understand exactly what a deep neural network
has learned, recently some progress has been made.
DeepMind's \emph{Psychlab} uses tests from psychology implemented in a
3D environment to understand deep RL agents.
The tests led them to a simple improvement of the UNREAL agent \citep{Leibo2018}.
\citet{Zahavy2016} instead use the dimensionality reduction technique
t-SNE on the activations of the top neural network
layer in DQN \citep{dqn}, revealing how DQN represents policies in ATARI games.
Looking beyond RL, \citet{Olah2017} summarize work on visualization of
features in image classification networks in a beautiful Distill post.
Another line of work tries to explain what text and speech networks have
learned \citep{Alvarez-Melis2017,Belinkov2017,Lei2016}.

\subsection{Safe learning}

During training, a standard Deep RL agent
such as DQN commits on the order of a million catastrophic mistakes such
as jumping off a cliff and dying \citep{Saunders2017}.
Such mistakes could be very expensive if they happened in the real world.
Further, we do not want an AGI to accidentally set off all nuclear weapons
in a burst of curiosity late in its training phase.
\citet{Saunders2017} propose to fix this by training a neural network to
detect potentially catastrophic actions from training examples provided by
a human.
The catastrophe detector can then override the agent's actions whenever
it judges an action to be too dangerous.
Using this technique, they manage to avoid all catastrophes in simple settings,
and a significant fraction in more complex environments.
A similar ideas was explored by \citet{Lipton2016}.
Instead of using human-generated labels,
their catastrophe detector was trained automatically on the agent's
catastrophes.
Unsurprisingly, this reduces but does not eliminate catastrophic mistakes.
A survey over previous work on safe exploration in RL is provided by
\citep{Garcia2015}.

\subsection{Other}

\paragraph{Boxing and oracles}
\citet{Armstrong2012oracles,Armstrong2017oracles} argue that oracles that
only answer questions may be more safe than other types of AGI.
\citet{Sarma2017} suggest that computer algebra systems are concrete
instances of oracles suitable for further study.
Essentially any type of AGI can be turned into an oracle by constraining its interaction
with the real world.
While it may be infeasible to keep a highly intelligent AGI from
breaking out of its ``box'' \citep{Yudkowsky2002,Alfonseca2016},
this may still be a useful technique for constraining more limited AGIs
\citep{Babcock2017}.

One way to box an AGI is to homomorphically encrypt it.
\citet{Trask2017} shows how to train homomorphically encrypted neural networks.
By homomorphically encrypting an AGI, its predictions and actions also come out encrypted.
A human operator with the secret key can choose to decrypt them only
when he wants to.

\paragraph{Tripwires}
\citet{Martin2016} use the AIXI framework to show that AGIs with rewards
bounded to a negative range (such as $[-1, 0]$) will
prefer their subjective environment to end.
In the AIXI framework, ``death'' always has an implicit reward of 0.
This may lead such systems to self-destruct once they are intelligent
enough to figure out how to do so.
Thus, a negative reward range may be used as a type of \emph{tripwire}
or \emph{honeypot}
\citep[Ch.~9]{Bostrom2014}, preventing superintelligent AGI before we are ready for it.

\paragraph{Meta-cognition}
\posscite{Garrabrant2016,Garrabrant2017} theory of logical induction leads to
several theorems about systems reasoning about their own computations in a
consistent manner, avoiding Gödelian and Löbian obstacles.
\citet{Fallenstein2015hol} shows that systems of higher-order logic can learn to
trust their own theorems, in a certain sense.

\paragraph{Weakening the agent concept}
\citet{Hedden2015} makes some progress on defining rational agency in cases
where there is no clear entity making the decisions, shedding some light on
the connection between personal identity and rational agency.


\section{Public Policy on AGI}
\label{sec:public-policy}

\paragraph{Recommendations}

In a collaboration spanning 14 
organizations, \citet{Brundage2018} consider scenarios for how AI and AGI
may be misused and give advice for both policy makers and researchers.
Regulation of AI remains a controversial topic, however.
On the one hand, \citet{Erdelyi2018} call for global regulatory body.
Others worry that regulations may limit the positive gains from AI
\citep{Gurkaynak2016,Nota2015}, and recommend
increased public funding for safety research \citep{Nota2015}.
\citet{Baum2017promotion} is also wary of regulation, but for slightly
different reasons.
He argues that \emph{extrinsic} measures such as regulations run the risk
of backfiring, making AI researchers look for ways around the regulations.
He argues that \emph{intrinsic} measures that make AI researchers want
to build safe AI are more likely to be effective,
and recommends either purely intrinsic methods or combinations of intrinsic
and extrinsic approaches.
He lists some ideas for intrinsic approaches:
\begin{itemize}
\item
  Setting social norms and contexts for building safe AI, by creating and
  expanding groups of AI researchers that openly promote safe AI, and by having
  conferences and meetings with this agenda.
\item
  Using ``allies'' such as car
  manufacturers (and the military) that want safe AI.
  AI researchers want to satisfy these organizations in order to access
  their funding and research opportunities.
\item Framing AGI less as winner-take-all race,
  as a race implies a winner.
  Instead, it is important to emphasize that the result of a speedy race
  where safety is de-prioritized is likely to be a very unsafe AGI.
  Also, framing AI researchers as people that are good (but sometimes do bad
  things), and emphasize that their jobs are not threatened by a focus on
  safety.
\item
  Making researchers publicly state that they care about safe AI
  and then reminding them when they don't follow it can lead to cognitive
  dissonance, which may cause researchers to believe they want safe AI.
\end{itemize}
\citet{Armstrong2016race} counterintuitively
find that information sharing between teams developing AGI exacerbates the risk
of an AGI race.

\paragraph{Policy makers}
Although public policy making
is often viewed as the domain of public bodies, it should be remembered that
many organizations such as corporations, universities and NGOs frequently become
involved through advocacy, consulting, and joint projects.
Indeed, such involvement can often extend to de facto or ``private'' regulation via
organizational guidelines, organizational policies, technical standards and
similar instruments.

Professional organizations have already taken a leading role.
The IEEE, for example, is developing guidelines on Ethically Aligned Design
\citep{IEEE2017,IEEE2017a}.
Meanwhile, the ACM and the SIGAI group of AAAI have co-operated to establish a
new joint conference on AI, ethics and society, AIES \citep{AIMatters2017}.
Economic policy and technical standards organizations have also started to engage: for
example, the OECD has established a conference on ``smart policy making'' around AI
developments \citep{OECD2017} and ISO/IEC has established a technical committee on AI
standards \citep{ISO/IEC2017}.
Corporations and corporate consortia are also involved,
typically through the public-facing aspects of their own corporate policies
\citep{Intel2017,IBM2018} or through joint development of safety policies and
recommendations which
consortia members will adopt \citep{Partnership2016}.

Finally, in addition to the traditional public roles of academia and academics, there are an
increasing number of academically affiliated or staffed AI organizations.
With varying degrees of specificity, these work on technical, economic,
social and philosophical aspects of AI and AGI.
Organizations include
the Future of Humanity Institute (FHI),
the Machine Intelligence Research Institute (MIRI),
the Centre for the Study of Existential Risk (CSER) and
the Future of Life Institute (FLI).

\paragraph{Current policy anatomy}

It could be said that public
policy on AGI does not exist.
More specifically, although work such as \citet{Baum2017projectsurvey}
highlights the extent to which AGI is a distinct endeavor with its own
identifiable risk, safety, ethics (RISE) issues,
public policy AGI is currently seldom separable from default public policy on AI taken as
a whole (PPAI).
Existing PPAI are typically structured around
(a) significant financial incentives (e.g. grants, public-private
co-funding initiatives, tax concessions) and
(b) preliminary coverage of ethical, legal and social issues (ELSI)
with a view to more detailed policy and legislative proposals later on
\citep{Miller2018,FTI2018}.

In the case of the EU, for example, in addition to experimental regulation
with its new algorithmic decision-making transparency requirements in
\citep[Article 22, General Data Protection Regulation]{EURlex2016}
, its various bodies and their industry partners
have committed over 3 billion Euro to AI and robotics R$\&$D and
engaged in two rounds of public consultation on the European Parliament’s
proposed civil law liability framework for AI and robotics \citep{Ansip2018}.
However, the much demanded first draft of an overarching policy framework
is still missing, being slated for delivery by the
European Commission no earlier than April 2018.

Elsewhere, spurred into action by the implications of the AlphaGo
victory and China’s recent activities (outlined below), South Korea and Japan have already
rapidly commenced significant public and public-private investment
programs together with closer co-ordination of state bodies, industry and
academia \citep{Ha2016,Volodzsko2017}.
Japan is also additionally allowing experimental
regulation in some economic sectors
\citep{Takenaka2017}.
The UK has started work on a preliminary national policy framework on robotics
and AI \citep{UKP2017,Hall2017},
and have established a national Centre for Data Ethics and Innovation
\citep{CSER2017}.

\paragraph{Current policy dynamics}
Although there is substantial
positive co-operation between universities, corporations and other
organizations,
there is a negative dynamic operating the nation-state, regional and
international context.
Contrary to the expert recommendations above, there is
increasing rhetoric around an AI ``arms race'' \citep{Cave2018},
typified by President Vladimir Putin’s September 2017 comment that
``... whoever becomes the leader in [the AI] sphere
will become the leader in the world'' \citep{Apps2017}.
Relatedly,
China’s 8 July 2017 AI policy announcement
included being the global leader in AI technology by 2030
\citep{PRC2017,Kania2018,Ding2018}.
It also included aims of ``creating a safer, more comfortable and convenient society.''
Alongside this policy shift has been increased Sino-American competition for AI
talent \citep{Cyranoski2018}.
In the US,
the Obama Administration began consultation and
other moves towards a federal policy framework for AI technology investment, development
and implementation
\citep{Agrawal2016,WhiteHouseOSTP2016}.
However, the Trump Administration abandoned the effort
to focus mainly on military spending on AI and cyber-security \citep{Metz2018}.

\paragraph{Policy outlook}
Given the above, looking forward it would appear that
the organizations noted above will have to work hard to moderate the negative
dynamic currently operating at the nation-state, regional and international level.
Useful guidance for researchers and others engaging with public policy
and regulatory questions on AI is given by \citet{Eightythousand2017}.
Further references on public policy on AGI can be found in \citep{Sotala2014,Dafoe2017}.

\section{Conclusions}
\label{sec:litrev-conclusions}

AGI promises to be a major event for humankind.
Recent research has made important progress on how to think about potential
future AGIs, which enables us to anticipate and (hopefully) mitigate problems
before they occur.
This may be crucial, especially if the creation of a first AGI leads to an
``intelligence explosion''.
Solutions to safety issues often have more near-term benefits as well,
which further adds to the value of AGI safety research.

It is our hope that this summary will help new researchers enter the field
of AGI safety, and provide traditional AI researchers with an overview of
challenges and design ideas considered by the AGI safety community.

\printbibliography

\end{document}